\documentclass{article}

\usepackage[preprint]{corl_2024} 

\usepackage[dvipsnames,table]{xcolor}
\usepackage{graphicx}
\usepackage{booktabs}
\usepackage{amsmath}
\usepackage{multirow}
\usepackage{makecell}
\usepackage{csquotes}
\usepackage{wrapfig}
\usepackage{diagbox}
\usepackage{caption} 
\usepackage{bbm}
\usepackage{algorithm}
\usepackage{algpseudocode}
\usepackage{}
\definecolor{tablegray}{RGB}{235,235,235}
\definecolor{figblue}{RGB}{80, 219, 224}
\definecolor{figgreen}{RGB}{183, 255, 95}

\newcommand{\ie}{\textit{i.e. }}

\newcommand\blfootnote[1]{
    \begingroup
    \renewcommand\thefootnote{}\footnote{#1}
    \addtocounter{footnote}{-1}
    \endgroup
}

\title{\includegraphics[height=1.5em]{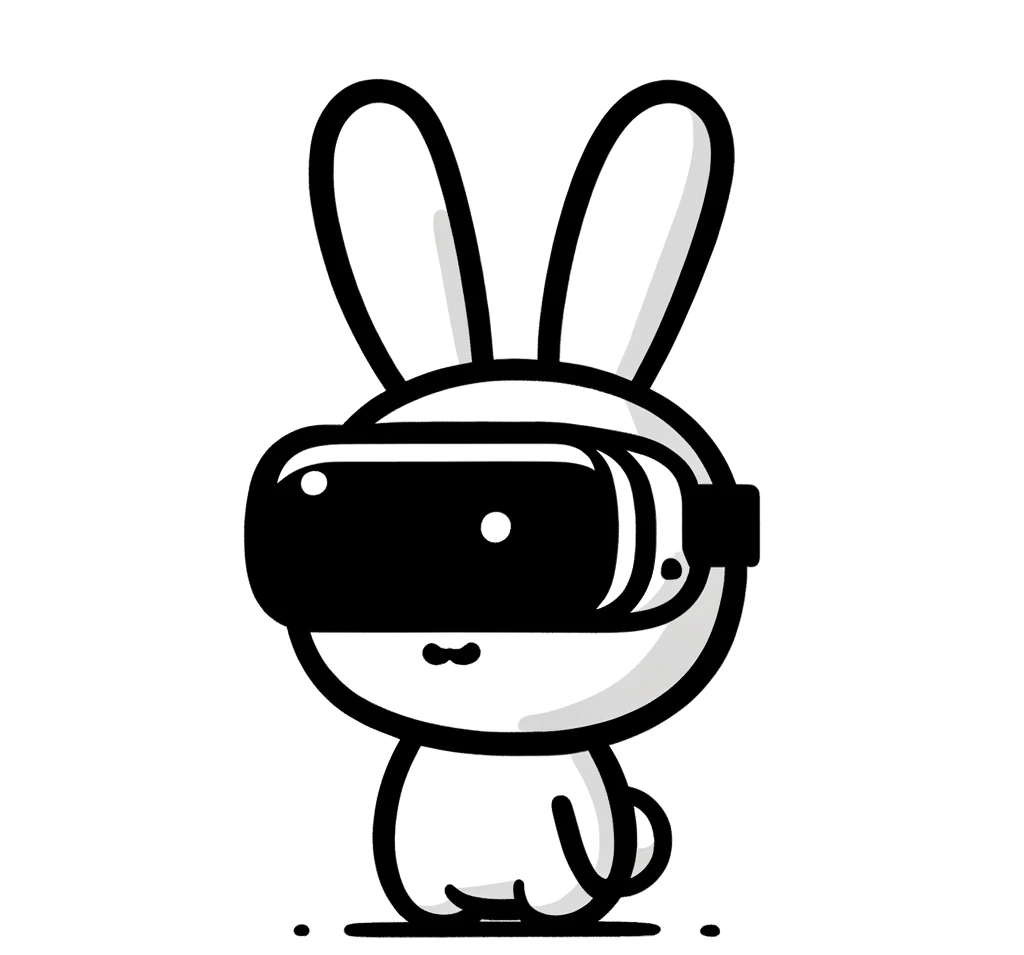}Bunny-VisionPro: Real-Time Bimanual Dexterous Teleoperation for Imitation Learning}

\author{
  Runyu Ding$^{1*}$, Yuzhe Qin$^{2*}$, Jiyue Zhu$^{2*}$, Chengzhe Jia$^2$, \\ \textbf{Shiqi Yang$^2$, Ruihan Yang$^2$, Xiaojuan Qi$^1$, Xiaolong Wang$^2$}\\
  $^1$ The University of Hong Kong,~~ $^2$ University of California, San Diego\\
}

\begin{document}
\maketitle


\vspace{-3em}
\begin{center}
    \href{https://dingry.github.io/projects/bunny\_visionpro}{https://dingry.github.io/projects/bunny\_visionpro}
\end{center}
\vspace{-0.5em}
\begin{abstract}
Teleoperation is a crucial tool for collecting human demonstrations, but controlling robots with bimanual dexterous hands remains a challenge. Existing teleoperation systems struggle to handle the complexity of coordinating two hands for intricate manipulations. We introduce Bunny-VisionPro, a real-time bimanual dexterous teleoperation system that leverages a VR headset. Unlike previous vision-based teleoperation systems, we design novel low-cost devices to provide haptic feedback to the operator, enhancing immersion. Our system prioritizes safety by incorporating collision and singularity avoidance while maintaining real-time performance through innovative designs.
Bunny-VisionPro outperforms prior systems on a standard task suite, achieving higher success rates and reduced task completion times. Moreover, the high-quality teleoperation demonstrations improve downstream imitation learning performance, leading to better generalizability. Notably, Bunny-VisionPro enables imitation learning with challenging multi-stage, long-horizon dexterous manipulation tasks, which have rarely been addressed in previous work.
Our system's ability to handle bimanual manipulations while prioritizing safety and real-time performance makes it a powerful tool for advancing dexterous manipulation and imitation learning.
\end{abstract}


{\blfootnote{{Runyu Ding was an intern at UCSD during the project}}}

\begin{figure}[h!]
    \centering
    \vspace{-1.5em}
    \includegraphics[width=\linewidth]{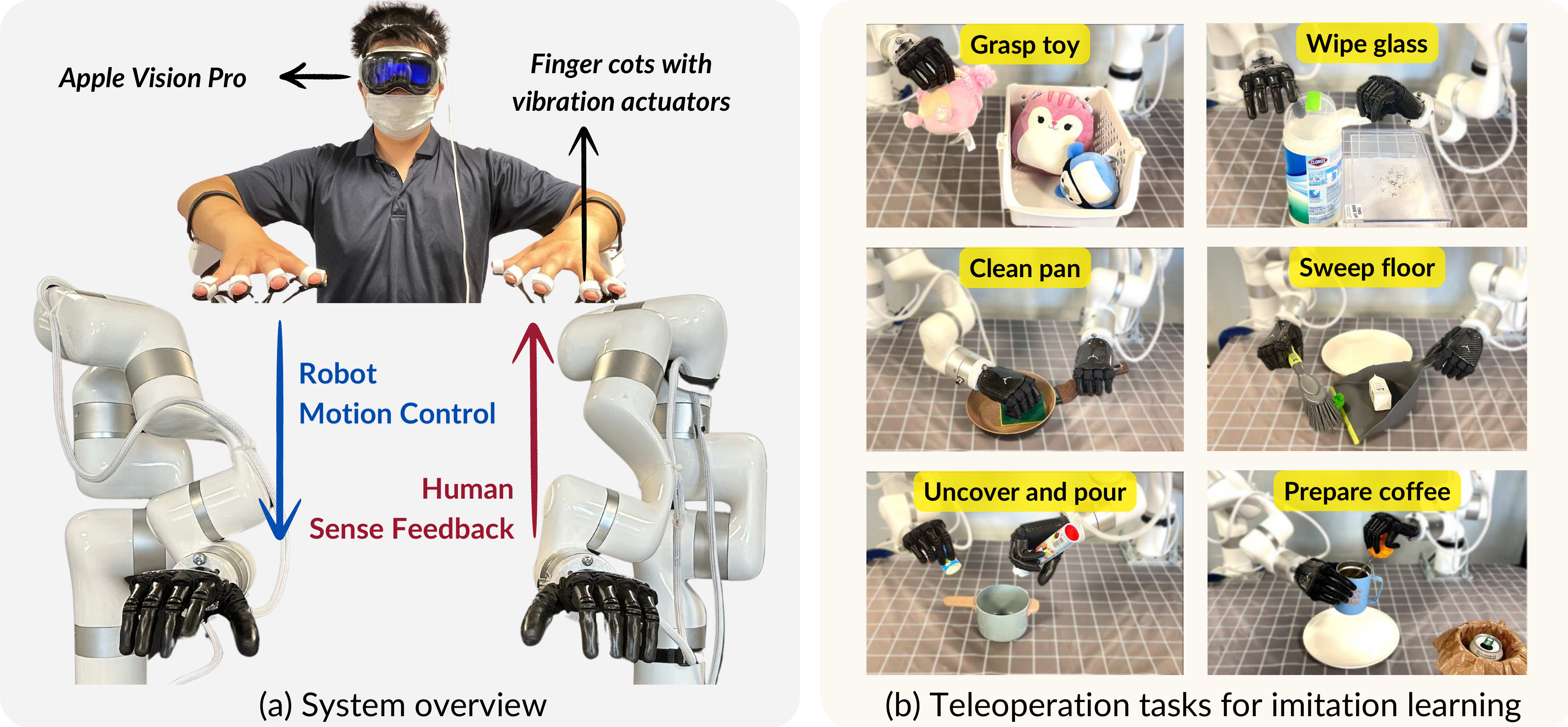}\label{fig:teleop}
    \caption{\textbf{System Overview and Task Suits.} \textbf{(a)} Hand poses captured by Apple Vision Pro are converted into robot motion control commands for real-time teleoperation. The robot provides sensory feedback, including vision and touch, to operators via Vision Pro and actuator-equipped finger cots. \textbf{(b)} We design diverse short-horizon (left-column) and long-horizon tasks (right-column) to evaluate teleoperation performance and its application for imitation learning.}
    \label{fig:teaser}
    \vspace{-0.5em}
\end{figure}

\newcommand{\xjqi}[1]{\textcolor[rgb]{1,0.0,0.0}{{{[xjqi:#1]}}}}

\section{Introduction}
\vspace{-5pt}
Playing a Virtual Reality (VR) game is an immersive and intuitive experience where your hand and arm movements seamlessly translate into the action of a virtual character. Now, imagine controlling a bimanual robot in the real world with the same ease: operators use their own movements to guide the robot's motion, just as they would in a VR game. This paradigm shift in robot teleoperation opens up exciting possibilities for more intuitive and accessible human-robot interaction. Recent advancements in VR technology, such as the Apple Vision Pro, have made this concept a possibility.

Yet, translating this concept into a practical teleoperation system presents significant challenges due to the intricate motion required to perform human-like manipulations. The complexity is further amplified for high Degree-of-Freedom (DoF) hand-arm systems, where the operator must coordinate two arms and hands to execute tasks requiring spatiotemporal synchronization. Achieving responsive control is crucial, as delays can lead to imprecise robot motion~\citep{hauser2013recognition}. Besides, ensuring safety by mitigating risks such as collision and singularity adds another layer of complexity~\citep{zhong2024attentiveness}.

In this paper, we introduce Bunny-VisionPro, a novel real-time bimanual dexterous teleoperation system. As illustrated in Fig.~\ref{fig:teaser}, our system utilizes Vision Pro's tracking capabilities to capture operator hand movements and translate these movements into precise robotic commands. It addresses both arm and dexterous hand control, prioritizing safety and real-time performance.
Crucially, Bunny-VisionPro features three innovative modules:
(i) The arm motion control module, which can handle robot singularity and collision avoidance in real time without the need for high-end GPUs
This module ensures the safe and smooth operation of robot arms, even in complex bimanual manipulation scenarios.
(ii) The dexterous hand retargeting module, which accurately maps human finger movements to the robotic hand. This module features its unique ability to perform retargeting for robots with loop joints, such as four-bar linkages, in real time. 
(iii) The haptic feedback module, which uses low-cost Eccentric Rotating Mass (ERM) actuators (\$1.2 each) to provide tactile sensations. This novel design, paired with a VR headset, offers a more immersive and realistic experience as the robot interacts with its environment, creating a sensation that the operator is the robot itself.

The advanced capabilities of Bunny-VisionPro enable the collection of high-quality demonstrations for dexterous, bimanual, and long-horizon imitation learning tasks. By leveraging the intuitive and responsive teleoperation system, operators can perform intricate manipulations and collect diverse data for learning algorithms, enabling them to learn and generalize to new scenarios effectively. 

We evaluate our Bunny-VisionPro system on the Telekinesis~\citep{sivakumar2022robotic} benchmark, achieving 11\% higher success rates and reducing task completion time by 45\% compared to prior systems. These enhancements are especially pronounced in multi-stage, long-horizon tasks.
Imitation learning policies trained on demonstrations collected by our system show a 20\% improvement in generalization on novel poses and unseen objects compared to using data from previous work~\citep{qin2023anyteleop}. These results demonstrate the system's superior performance in executing complex bimanual manipulation tasks and its effectiveness in collecting high-quality data.

\vspace{-5pt}
\section{Related Work}
\label{sec:related}
\vspace{-5pt}
\textbf{Teleoperation with Gripper.}
Classical teleoperation methods can be categorized into two main approaches based on their control objectives. The first approach, exemplified by ALOHA~\citep{zhao2023learning, fu2024mobile, fang2023low}, uses joint-space mapping~\citep{toedtheide2023force} within leader-follower setups. Although enabling impressive bimanual manipulation, this approach is robot-specific, requiring kinematic equivalence between the leader and follower robots~\citep{wu2023gello}. It also places the burden of managing collision and singularity on the human operator. The second approach prioritizes end-effector control~\citep{roboturk-bimanual}, with arm joint positions calculated using inverse kinematics. Various input devices, such as motion capture systems~\citep{zhao2012combining, liu2021semi}, inertia sensors~\citep{zhang2018feasibility}, and VR controllers~\citep{zhang2018deep, helping-robot, pan2021augmented, iyer2024open} have been utilized. Nevertheless, these systems often employ simple 1 or 2 DoF grippers, which limits their dexterity. In contrast, our work aims to teleoperate high-DoF hand-arm systems for complex manipulation tasks.

\textbf{Dexterous Teleoperation.}
Teleoperating dexterous hands is challenging due to their high DoF and complex kinematics. Glove-based systems~\citep{wang2024dexcap, liu2019high, mosbach:humanoids2022, schwarz2021nimbro}, like the MANUS glove~\citep{manus} used by Tesla Bot, can track the operator's finger movements but are costly and require specific hand sizes. Recent vision-based approaches, such as AnyTeleop~\citep{qin2023anyteleop}, facilitate dexterous hand-arm teleoperation using cameras~\citep{sivakumar2022robotic, dexpilot, li2022dexterous, from-one-hand} or VR headsets~\citep{holodex, yim2022wfh, ponomareva2021grasplook}. However, AnyTeleop requires sophisticated GPU processing for arm motion computation and is primarily designed for a single arm. A concurrent work~\citep{lin2024learning} controls the Ability hand with reduced DoF using grip buttons, sacrificing dexterous finger gaiting to avoid retargeting latency. In contrast, our retargeting modules can handle the Ability hand's four-bar linkage structure at 300Hz with one CPU core, enabling full dexterity. Our system also integrates collision and singularity avoidance while maintaining real-time performance.

\textbf{Imitation Learning from Demonstration.} Imitation learning enables robots to mimic human behaviors through expert guidance. Pioneering research utilizing deep learning~\citep{bc-z,rt-1,rt22023arxiv,reed2022a} develop policies to generate robot control commands based on image~\citep{zhao2023learning,chi2023diffusionpolicy,pari2021surprising,robo-mimic,haldar2023teach,wang2023mimicplay} and point cloud data~\citep{shridhar2022peract,Ze2024DP3,goyal2023rvt,gervet2023actd}, and further marking progress in bimanual systems~\citep{wang2024dexcap,zhao2023learning,fu2024mobile,grannen2023stabilize}. Additionally, recent works incorporate tactile data~\citep{yang2023seq2seq,lin2024learning} to enrich the sensory data pool for robot learning.
Given that demonstration collection is labor-intensive yet highly critical for effective imitation learning, our system offers a compelling solution that streamlines the collection of high-quality bimanual dexterous data, greatly enhancing accuracy and generalization across various learning algorithms.
\vspace{-5pt}
\section{Teleoperation System}
\label{sec:teleop}
\vspace{-5pt}

\subsection{Overview}
\label{sec:teleop_overview}
\vspace{-5pt}
\begin{figure}
    \vspace{-15pt}
    \centering
    \includegraphics[width=\linewidth]{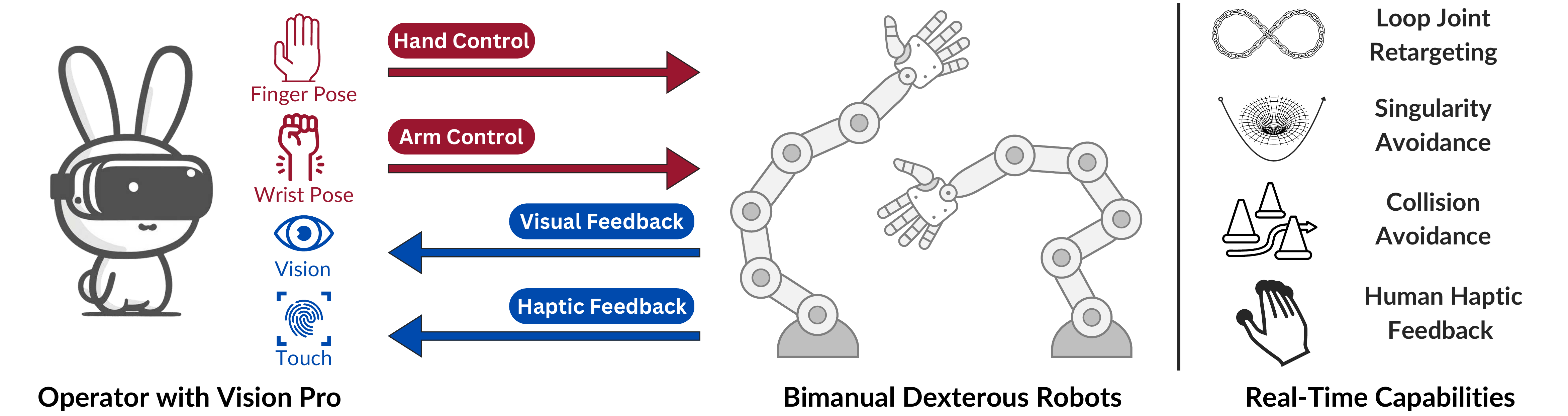}
    \caption{\textbf{Teleoperation System.} The operator controls the robot hand and arm using finger and wrist poses, respectively. The system's visual and haptic feedback, combined with its four real-time capabilities, provides an intuitive and immersive VR experience for the operator.}
    \label{fig:pipeline}
    \vspace{-2em}
\end{figure}

Bunny-VisionPro is a modular teleoperation system using a VR headset's hand and wrist tracking capabilities to control a high-DoF bimanual robot, as shown in Fig.~\ref{fig:pipeline}. The system consists of three decoupled components: hand motion retargeting, arm motion control, and human haptic feedback.

The hand motion retargeting module maps the operator's finger poses to the robot's dexterous hands, enabling intuitive dexterous manipulation. Concurrently, the arm motion control module uses the operator's wrist poses as input to compute joint angles for the robot arms while considering collision avoidance and singularity handling. The human haptic feedback module converts the tactile readings from sensors mounted on the robot's hands into actuation signals for the wearable Eccentric Rotating Mass (ERM) actuators on the operator's hands. This provides the operator with real-time haptic feedback, enhancing their sense of presence and allowing for more precise manipulation. 

Coordinating bimanual robots during teleoperation also requires maintaining the distance between two robot hands to match the distance between two operators' hands, ensuring natural, coordinated movements. However, the initial distance between the robot hands may not match that of the human hands. To address this, we design several initialization modes that align the robot's hands with the operator's hands on the fly when the robot starts to move. Different tasks may benefit from different initialization modes. This step creates a consistent starting point, enabling intuitive and efficient bimanual task execution throughout the teleoperation process. More details are in Appendix.

For communication, we stream hand pose results from Vision Pro to the computer via~\citep{Park_Teleopeation_System_using}. The modular architecture allows for better extensibility and enables each module to run in a separate computation process, preventing latency accumulation in the system and ensuring real-time control.

\subsection{Robot Hand Motion Retargeting}
\label{sec:teleop_hand}
\vspace{-3pt}
The hand motion retargeting module translates human finger poses into corresponding robot joint positions. This is achieved by minimizing the difference between the fingertip keypoint vectors~\cite{dexmv} of the human hand and the robotic hand, formulated as an online optimization problem:

\vspace{-10pt}
\begin{equation}
\label{eq:retargeting_main}
L_\text{hand} = \sum_{i=1}^{N} || \alpha v_i - \text{FK}_i(q) ||^2_2 + \beta||\Delta q||^2_2 \quad \quad \mathrm{s.t.} \quad q^{l} \le q  \le q^{u}. 
\end{equation}
The objective function $L_\text{hand}$ consists of two terms: (i). The first term minimizes the distance between the scaled human hand keypoint vectors $\alpha v_i$ and the robot hand keypoint vectors computed by forward kinematics function $\text{FK}_i(q)$, where $v_i$ represents the $i^\text{th}$ fingertip keypoint vector, $N$ is the total number of vectors, $||\cdot||_2$ denotes Euclidean norm, $\alpha$ is the scaling factor to adjust for size differences between the human and robot hand and $q$ is the joint positions of the robot hand bounded by the lower and upper limit, $q^l$ and $q^u$. (ii). The second term enforces temporal smoothness by penalizing large joint position changes $\Delta q$ between consecutive frames, with weight $\beta$. The optimization variable is $q$ and the objective is solved using Sequential Quadratic Programming (SQP)~\citep{boggs1995sequential, johnson2021nlopt}, a gradient-based method that iteratively solves quadratic approximations of the problem.

Our work uniquely addresses the real-time handling of loop joints, common in dexterous hands, where the number of joints exceeds the actual DoF. For a robot with $n$ joints, the positions of $n-k$ passive joints depend on the other $k$ active joint, represented as:
\vspace{-5pt}
\begin{equation}
    q_j=c_j(q_1, q_2, ..., q_k), ~\text{for}~ j \in \{k+1, \dots, n\},
\end{equation}\label{eq:retargeting_passive_joint}where the passive joint $j$'s position $q_j$ is not independent but constrained by function $c_{j}$. $n$ is the total number of joint positions.
Instead of adding equality constraints to the optimization problem in Eq.~\eqref{eq:retargeting_main}, which can lead to instability and increased computation time due to the highly non-linear forward kinematic function $\text{FK}(q)$, we reformulate the problem into a reduced dimension where the optimization variable has $k$ dimensions (active joints) instead of $n$ (total joints). Specifically, in the forward pass of optimization, the positions of passive joints are computed as Eq.~\eqref{eq:retargeting_passive_joint}. 
In the backward pass, the gradients of the passive joints are backpropagated to the active joints and added to their original gradients. Thereby, the gradient for the active joint $i$ is given by:

\vspace{-5pt}
\begin{equation}
\text{grad}_{i} = \frac{\partial L_\text{hand}}{\partial q_i}+ \sum_{j=k+1}^{n} \frac{\partial c_j}{\partial q_i} \frac{\partial L_\text{hand}}{\partial q_j}.
\end{equation}
This approach eliminates the need for SQP to handle affine approximations of the constraints, making the problem more tractable. Experimental results in Tab.~\ref{tab:profile} show a \textbf{10.2x speedup} when using this method to solve the four-bar linkage structure (a type of loop joint) in the Ability Robot Hand.

\subsection{Robot Arm Motion Control}
\label{sec:teleop_arm}
\vspace{-3pt}
Arm motion control involves computing the joint trajectories of the robot arm based on the wrist pose detected by Vision Pro. Traditional approaches often rely on closed-form solutions for Inverse Kinematics (IK) and handle constraints such as singularity by exploring the null space~\citep{dietrich2012reactive}. However, for typical 7-DoF arms, the 1-DoF null space may not provide enough flexibility. Modern approaches formulate the problem as an optimization task~\citep{bhardwaj2022storm}, similar to retargeting. It relaxes the rigid constraints of IK to accommodate additional factors.
Inspired by these works, we have developed a unified and efficient optimization objective $L_\text{arm}$ in Eq.\eqref{eq:arm_control_main} that integrates IK, collision avoidance, and singularity management for real-time motion control. 

\vspace{-5pt}
\begin{equation}    L_\text{arm}=L_\text{ik}+L_\text{sin}+L_\text{col},
\end{equation}\label{eq:arm_control_main}
The first term concerns the IK objective:
\begin{equation}
L_\text{ik}=\beta_\text{pos}||p_\text{ee}-p_\text{wrist}||_2 + \beta_{\text{rot}}\cdot\arccos(2 \langle q_\text{ee}, q_\text{wrist}\rangle^2 - 1). 
\end{equation}
Here, $p_\text{ee}$ and $p_\text{wrist}$ represent the robot end effector and human wrist position, respectively, while $q_\text{ee}$ and $q_\text{wrist}$ are their corresponding quaternions. $\langle \cdot, \cdot \rangle$ means inner product.
The weights $\beta_\text{pos}$ and $\beta_\text{rot}$ allow for balancing the errors between position and rotation.
Additionally, the objectives for singularity and collision avoidance are as follows:

\vspace{-5pt}
\begin{equation}
\begin{small}
    \quad L_\text{sin}=
    \left\{ 
        \begin{array}{rcl}
        1-\frac{\sqrt{\det[\mathbf{JJ^T}]}}{\lambda}, & \mbox{if} \mkern9mu s_0<s_\text{low} \\
        0, & \mbox{otherwise}
        \end{array},
    \right.\quad
    L_\text{col}=\frac{1}{\sum_{i=1}^m \sum_{j=1}^m \mbox{dist}(e_i,e_j)\mathbbm{1}(e_i,e_j) + \epsilon}.
\end{small}
\end{equation}
In the singularity avoidance objective $L_\text{sin}$, $\mathbf{J}$ represents the spatial Jacobian matrix, $s_0$ is the smallest singular value, $s_\text{low}$ is a conditional trigger for penalizing singularity, and $\lambda$ acts as a temperature factor. Theoretically, both $s_0$ and the manipulability index $\sqrt{\det[\mathbf{JJ^T}]}$ can measure singularity~\citep{yoshikawa1985manipulability}, and the robot is nearing singularity when either value is low. We opt for the manipulability index as the objective due to its suitability for gradient-based optimization while using $s_0$ as the condition. 
Thus, $L_\text{sin}$ only takes effect when the robot approaches singularity.

For collision detection, the GJK algorithm~\citep{gilbert1988fast} is often employed but can be time-intensive even with convex meshes. To efficiently compute the self-collision cost $L_\text{col}$, we model each robot link as a collection of $m$ spheres. The function $\text{dist}(\cdot, \cdot)$ calculates the Euclidean distance between spheres $e_i$ and $e_j$, while $\mathbbm{1}(\cdot, \cdot)$ serves as an indicator function to evaluate the necessity of collision checking, with 
$\epsilon$ provides numerical stability. Collisions are disregarded between spheres within the same link or between spheres of a parent and its direct child. 
This simplification not only accelerates computation but also renders the distance function differentiable. 

\vspace{-3pt}
\subsection{Human Haptic Feedback Device}\label{sec:haptic_feedback}
\begin{figure}
    \centering
    \vspace{-5pt}
    \includegraphics[width=\linewidth]{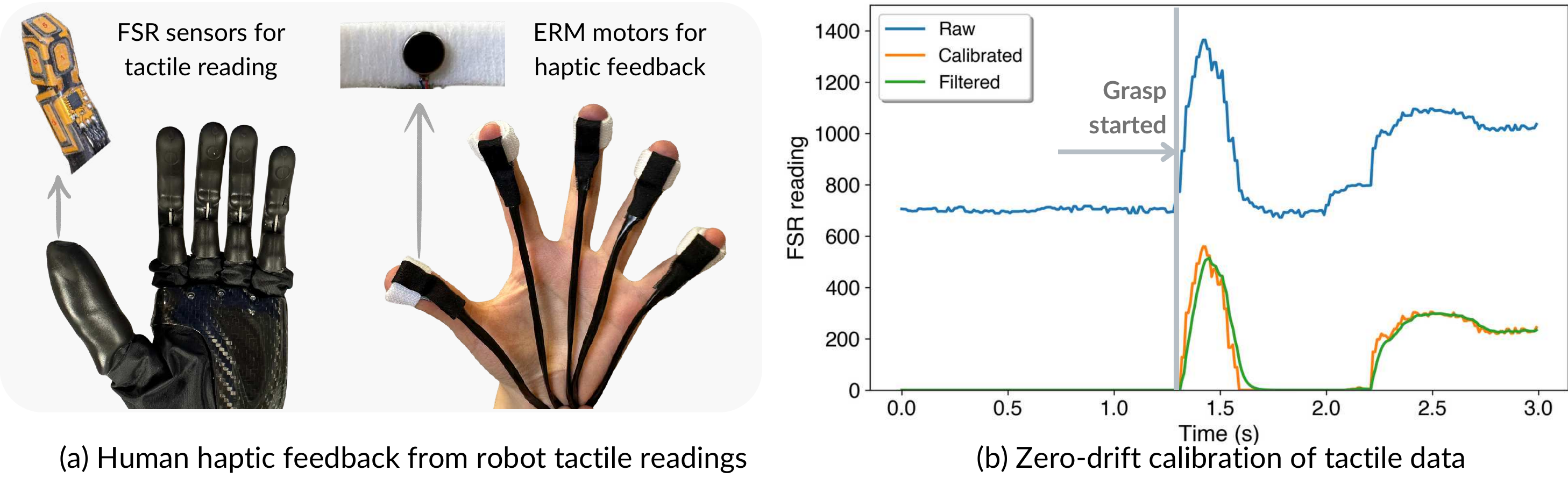}
    \caption{\textbf{Human Haptics Feedback Device.} Tactile readings from FSR sensors located in the robot's fingertips undergo calibration and low-pass filtering. The processed tactile signals then drive ERM motors to deliver touch feedback to human.}
    \label{fig:haptics}
    \vspace{-10pt}
\end{figure}

\vspace{-3pt}
Effective human manipulation relies on the integration of both visual and tactile feedback. However, many vision-based teleoperation systems~\cite{qin2023anyteleop,dexpilot,holodex,ponomareva2021grasplook, sivakumar2022robotic} neglect haptic feedback. To address this limitation, we have developed a cost-effective haptic feedback system using Eccentric Rotating Mass (ERM) actuators (Fig.~\ref{fig:haptics}(a)). This system first processes tactile signals from robot hands (Step I) and then drives vibration motors to simulate tactile sensations (Step II). Despite its low cost, our system enables the operator to perceive and respond to the environment more intuitively with a more immersive experience, leading to improved manipulation performance.

\noindent\textbf{Step I: Tactile Signal Processing.}
The Ability hand (Fig.\ref{fig:haptics}(a)) uses Force-Sensitive Resistors (FSRs) to measure finger pressure. However, FSR sensors suffer from imprecision and zero-drift problems\citep{barnea2012force, lakshmi2020calibration}, which are aggravated by deformable wrapping materials. To address this, we perform zero-drift calibration by recording baseline FSR readings at various joint positions and subtracting interpolated baseline values from real-time readings during operation. This efficient calibration can be performed autonomously each time teleoperation is initiated. Furthermore, we apply a low-pass filter to reduce noise and smooth the tactile data. As shown in Fig.~\ref{fig:haptics}(b), these signal-processing techniques significantly improve the quality and reliability of the tactile feedback. 

\noindent\textbf{Step II: Vibration Motor Driving.}
In this step, we convert the processed tactile signals to ERM actuator vibrations using an ELEGOO UNO board. Since ERM motors require constant input voltage, we employ Pulse-Width Modulation (PWM) to simulate continuous haptic strength by controlling ERM vibration intensity through modulated pulse width. To further enhance the stability of the haptic signals and the system's robustness, we add a Bipolar Junction Transistor (BJT) between each ERM actuator and its corresponding PWM pin. This ensures that the haptic feedback of each ERM motor is individually adjusted, only activated during the pulse width, and resistant to circuit noise.

\vspace{-5pt}
\section{System Evaluation}\label{sec:system_evaluation}
\vspace{-3pt}
We validate our Bunny-VisionPro system through profiling analysis (Sec.~\ref{sec:profiling}), comparing it with advanced teleoperation systems (Sec.~\ref{sec:exp_teleop}) on the Telekinesis benchmark~\citep{sivakumar2022robotic} and custom dexterous manipulation tasks. Note that the human haptic feedback device was not used in the teleoperation comparison experiments because all demonstration collections were completed before the haptic system was designed. We evaluate the usability of haptic feedback in a user study involving untrained operators  (Sec.~\ref{sec:exp_haptics}).

\subsection{Profiling Results}\label{sec:profiling}
\vspace{-5pt}
We profile the performance of different modules from Sec.\ref{sec:teleop} under specific conditions (Tab.\ref{tab:profile}). Our method retargets hand motion on the Ability hand, handling loop joints at a speed comparable to retargeting without loop joints. For motion control, even with singularity and collision avoidance, our modules run in real-time at $>60$ Hz.

\begin{table}[b]
\vspace{-10pt}
\begin{minipage}[h]{0.39\linewidth}
\centering
\resizebox{1\columnwidth}{!}{
\begin{tabular}{lc}
       \toprule
       CPU &  i7-12700KF  \\
       \midrule
       Module &  Time (ms)  \\
       \midrule
       Retargeting (w/o loop joints) & $2.54$  \\
       Retargeting (loop joints as Cons.) & $34.98$   \\
       Retargeting (loop joints ours) & $3.43$ \\
       Motion Control (IK only) &  $0.74$  \\
       Motion Control (+ Coll.) &  $7.85$  \\
       Motion Control (+ Sing.) &  $10.42$  \\
       Motion Control (+ Coll. + Sing.) &  $15.93$  \\
       Haptic Feedback (PWM) &  $0.04$ \\
       \bottomrule  
    \end{tabular}
}
\caption{\textbf{Profiling Results.} We profile different modules at certain conditions.}
\label{tab:profile}
\end{minipage}
\hfill
\begin{minipage}[h]{0.57\linewidth}
\centering
\resizebox{1\columnwidth}{!}{
    \begin{tabular}{lccc}
    \toprule
    Task & Telekinesis~\citep{sivakumar2022robotic} & AnyTeleop~\cite{qin2023anyteleop} & Ours \\
    \midrule
    Pickup Box Object &9/10  &\textbf{10/10} & \textbf{10/10} \\
    Pickup Fabric Toy &9/10 &\textbf{10/10} & \textbf{10/10}\\
    Box Rotation &6/10 &6/10 & \textbf{9/10}\\
    Scissor Pickup &7/10 &\textbf{8/10} & 7/10 \\
    Cup Stack &6/10 &9/10 & \textbf{10/10} \\
    Two Cup Stacking  &3/10  &7/10 & \textbf{9/10} \\
    Pouring Cubes onto Plate &7/10 & 7/10 & \textbf{10/10}  \\
    Cup Into Plate  &8/10  &\textbf{10/10}  & \textbf{10/10} \\
    Open Drawer  &9/10   &\textbf{10/10} & \textbf{10/10} \\
    Open Drawer \& Pickup Object &6/10 &9/10 & \textbf{10/10}\\
    \bottomrule
    \end{tabular}
}
\caption{\textbf{Teleoperation Results in Telekinesis Benchmark.} 10 single-arm tasks are evaluated.
} 
\label{tab:teleop_10tasks}
\end{minipage}\\
\end{table}
\vspace{-3pt}
\subsection{Real Robot Teleoperation Experiments}\label{sec:exp_teleop}

\noindent\textbf{Task Suits.}
The Teleskinesis~\citep{sivakumar2022robotic} benchmark consists of ten single-arm-hand manipulation tasks, as listed in Tab.~\ref{tab:teleop_10tasks}. To further gauge our system's capability primarily for bimanual dexterous manipulation, we designed three short-horizon tasks -- \textit{grasping toy}, \textit{cleaning pan} and \textit{uncovering and pouring} -- and three long-horizon tasks -- \textit{wiping glass}, \textit{sweeping floor} and \textit{preparing coffee}, as shown in Fig.~\ref{fig:teaser}. These tasks, except for \textit{grasping toy}, necessitate fine-grained coordination between two hands.

\noindent\textbf{Real-world Robot Setup.}
Our bimanual dexterous system consists of two UFactory xArm-7 robotic arms, each equipped with a 6-DoF Ability hand, resulting in a 24-DoF system. Each hand incorporates 30 tactile sensors distributed across five fingertips. For demonstration collection, two RealSense L515 cameras are positioned at the front and the top of the robot’s workspace to capture adequate visual observations.

\noindent\textbf{Baselines}.
We assess our system on the Teleskinesis benchmark against Teleskinesis~\citep{sivakumar2022robotic} and AnyTeleop~\citep{qin2023anyteleop}, using the results reported in the original papers for comparison.
For our custom tasks, to ensure a fair comparison, we integrate AnyTeleop with our Vision-Pro-based hand tracking to create AnyTeleop+ and use an identical real-robot setup as our Bunny-VisionPro.

\noindent\textbf{Teleoperation Results}.
As shown in Tab.~\ref{tab:teleop_10tasks}, Bunny-VisionPro matches or surpasses baseline methods in 9 out of 10 tasks, evidencing its system precision and robustness for hand manipulation teleoperation. 
However, in the \textit{scissor pickup} task, our system underperforms slightly due to the limited DoF in the Ability hand, which makes it challenging to insert fingers into the scissor handles.

In self-designed manipulation tasks (Tab.~\ref{tab:teleop_ourtasks}), our system outperforms AnyTeleop+~\cite{qin2023anyteleop}, achieving 11\% higher success rate with 45\% task completion time. This performance confirms its precision in replicating human actions and its responsiveness and efficiency. Additionally, Bunny-VisionPro demonstrates enhanced stability and robustness, reducing episode lengths by 19\% with lower variability.
In contrast, AnyTeleop+ struggles with complex trajectories and large end-effector pose changes, often leading to aggressive joint movements and unpredictable, potentially unsafe robot control, as evidenced by a 43\% increase in arm joint position changes during tasks.
Besides, our system's advantages are particularly pronounced in bimanual long-horizon tasks, where it excels in real-time, fine-grained coordination between two hands, leading to a better teleoperation experience.

\begin{table}[]
\centering
\begin{small}
\setlength\tabcolsep{3pt}
\scalebox{0.93}{
\begin{tabular}{llcccccc}
    \toprule

    Task & Description & Arm & System & Success ($\uparrow$) & Time ($\downarrow$) & EpLen ($\downarrow$) & $\Delta \text{Qpos}$ ($\downarrow$) \\
    \midrule

    \multicolumn{8}{c}{\textit{\textbf{Short-horizon Tasks}}}\\
    
    \multirow{2}{*}{Grasp Toy} & \multirow{2}{*}{\makecell[l]{Grasp toy and place in box.}} & \multirow{2}{*}{single} & AnyTeleop+ & \textbf{50 / 54} & 28.9 & 145$\pm$24 & 0.77 \\
    & & & Ours & 50 / 55 & \textbf{24.3} & \textbf{142$\pm$21} & \textbf{0.35} \\
    
    \rowcolor{gray!10}
    & & & AnyTeleop+ & 50 / 78 & 74.6 & 282$\pm$23 & 0.98 \\
    \rowcolor{gray!10}
    \multirow{-2}{*}{Clean Pan} & \multirow{-2}{*}{\makecell[l]{Grasp panhandle and sponge \\ to clean the pan.}} & \multirow{-2}{*}{dual} & Ours & \textbf{50 / 58} & \textbf{30.4} & \textbf{186$\pm$17} & \textbf{0.58} \\
    
    \multirow{2}{*}{Uncover \& Pour} & \multirow{2}{*}{\makecell[l]{Grasp container, remove lid,\\ and pour into bowl.}} & \multirow{2}{*}{dual} & AnyTeleop+ & 50 / 77 & 63.2 & 189$\pm$21 & 1.26 \\
    & &  & Ours & \textbf{50 / 63} & \textbf{41.5} & \textbf{183$\pm$15} & \textbf{0.86} \\
    
    \multicolumn{8}{c}{\textit{\textbf{Long-horizon Tasks}}}\\

    \rowcolor{gray!10}
    & & & AnyTeleop+ & 30 / 53 & 69.0 & 517$\pm$51 & 1.44 \\
    \rowcolor{gray!10}
    \multirow{-2}{*}{Wipe Glass} & \multirow{-2}{*}{Pull out wipes to clean glass} & \multirow{-2}{*}{dual} & Ours & \textbf{30 / 41} & \textbf{32.2} & \textbf{360$\pm$35} & \textbf{0.68} \\
    
    \multirow{2}{*}{Sweep Floor} & \multirow{2}{*}{\makecell[l]{Sweep and dump trash \\ using dustpan and broom}} & \multirow{2}{*}{dual} & AnyTeleop+ & 30 / 61 & 129.4 & 786$\pm$118 & 1.99 \\
    & & & Ours & \textbf{30 / 40} & \textbf{45.4} & \textbf{570$\pm$77} & \textbf{1.30} \\

    \rowcolor{gray!10}
    & & & AnyTeleop+ & 30 / 52 & 95.3 & 705$\pm$42 & 1.87 \\
    \rowcolor{gray!10}
    \multirow{-2}{*}{Prepare Coffee} & \multirow{-2}{*}{\makecell[l]{Pour coffee into bottle, stir \\ and hand it over to human}} & \multirow{-2}{*}{dual} & Ours & \textbf{30 / 44} & \textbf{55.2} & \textbf{576$\pm$33} & \textbf{1.05} \\

    \bottomrule
\end{tabular}
}
\end{small}
    \caption{\textbf{Teleoperation System Evaluation.} We measure collection success rate (Success), episode length (EpLen), collection time in minutes (Time), and joint position changes of the arms($\Delta \text{Qpos}$).}
    \label{tab:teleop_ourtasks}
    \vspace{-20pt}
\end{table}

\vspace{-5pt}
\subsection{User Study of Haptic Feedback}\label{sec:exp_haptics}
\vspace{-5pt}
\begin{wrapfigure}{R}{7.5cm}
    \centering
    \vspace{-12pt}
    \includegraphics[width=\linewidth]{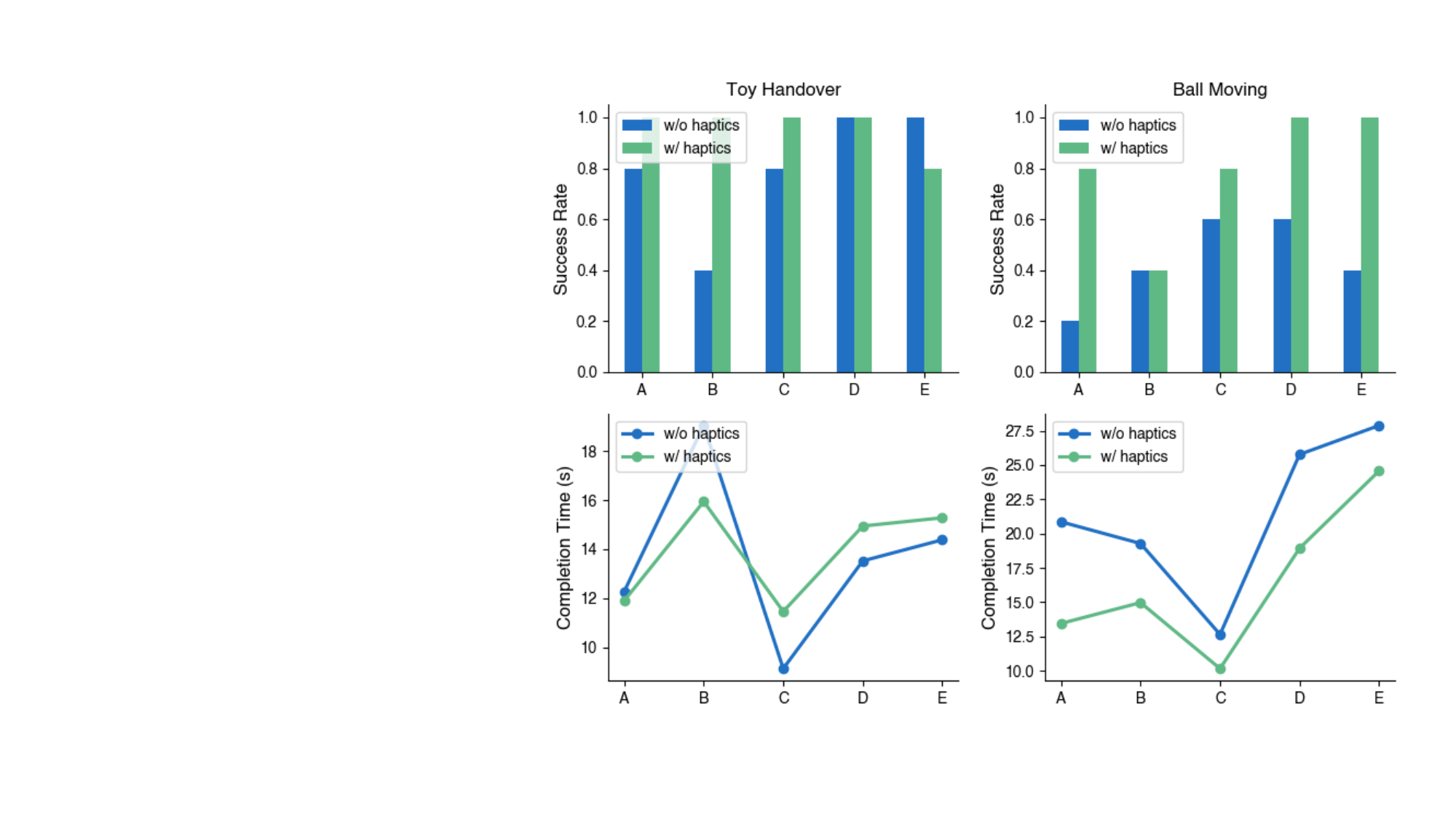}
    \caption{\textbf{User Study.} It evaluates the impact of haptic feedback on success rates and time efficiency.}
    \label{fig:user_study}
    \vspace{-10pt}
\end{wrapfigure}

We invite five untrained operators for a user study on human haptic feedback involving two tasks: (i) \textit{toy handover} and (ii) \textit{ball moving} (see Appendix for task figures). Each operator performs five trials per task, and we measure task success rates and average completion time. As shown in Fig.~\ref{fig:user_study}, haptic feedback maintains or improves success rates in 9 out of 10 comparisons. This boost is attributed to enhanced interaction awareness between the robot and the object, which prevents abnormal current generation in robot arms due to excessive pressure, thus promoting safer operations. 
Moreover, haptic feedback allows for quicker completion of the \textit{ball moving} task, which requires precise control due to the ball's low deformability. Operators report that the haptic feedback expedites object localization when their vision is partially obstructed and enhances their confidence in teleoperation.

\vspace{-5pt}
\section{Imitation Learning}
\vspace{-5pt}
To evaluate the quality of demonstrations collected by AnyTeleop+ and our system from an imitation learning perspective, we train several popular methods: ACT~\citep{zhao2023learning}, Diffusion Policy~\citep{chi2023diffusionpolicy}, and DP3~\citep{Ze2024DP3} and test their generalization performance of unseen scenarios. Besides, we investigate the effectiveness of tactile data in imitation learning. Although proven effective, the haptic feedback system is not used in the demonstration collection because it was designed after the data gathering was completed. Demonstration details are provided in Appendix.

\subsection{Main Results.}
\vspace{-5pt}
As shown in Tab.~\ref{tab:imitation_learning}, Bunny-VisoinPro significantly outperforms AnyTeleop+ by an average of 22\% success rate across three tasks using three different policies. This showcases the superior quality of the demonstrations collected by our system. Our system's effectiveness is particularly evident in bimanual tasks, \ie \textit{cleaning pan} and \textit{uncovering and pouring}. One possible reason for this improvement is that our system handles the alignment of bimanual robots delicately, enabling the collection of demonstrations with more natural robot motions.

\textbf{Generalization.} 
We assess the generalizability of policies learned on AnyTeleop+ and our demonstrations. It evaluates spatial generalization (SR-SG) with random object poses, and the ability to manipulate unseen objects (SR-U) varying in shape, size, and color. As shown in Tab.~\ref{tab:imitation_learning}, Bunny-VisionPro exceeds by 14\% SR-SG and 26\% SR-U, suggesting that high-quality, trajectory-consistent demonstrations significantly improve generalization in imitation learning. Details are in Appendix.

\textbf{Long-horizon Tasks.} 
For complicated, multi-stage tasks, the policy achieves notable success rates -- 73\% for sub-tasks and 38\% for entire tasks  -- with only 30 demonstrations. This manifests the reliability of long-horizon data collection of our system to serve advanced manipulation task learning.

\begin{table}[]
\centering
\begin{small}
\setlength\tabcolsep{5pt}
\scalebox{0.95}{
\begin{tabular}{lcccccccccc}
\toprule
\multirow{2}{*}{Task} & \multirow{2}{*}{System} & \multicolumn{3}{c}{ACT~\citep{zhao2023learning}} & \multicolumn{3}{c}{Diffusion Policy~\citep{chi2023diffusionpolicy}} & \multicolumn{3}{c}{DP3~\citep{Ze2024DP3}} \\
\cmidrule{3-11}
 &   & SR & SR-SG & SR-U & SR & SR-SG & SR-U & SR & SR-SG & SR-U \\
 \midrule

\multirow{2}{*}{Grasp toy} & AnyTeleop+ & \textbf{8/10} & \textbf{6/10} & \textbf{7/10} & 4/10 & 3/10 & 3/10 & \textbf{10/10} & 5/10 & 5/10\\
& Ours & \textbf{8/10} & 4/10 & 5/10 & \textbf{7/10} & \textbf{6/10} & \textbf{4/10} & \textbf{10/10} & \textbf{6/10} & \textbf{7/10} \\

\rowcolor{gray!10}
& AnyTeleop+ & 7/10 & 3/10 & 2/10 & 5/10 & \textbf{4/10} & 1/10 & 6/10 & 1/10 & 4/10 \\
\rowcolor{gray!10}
\multirow{-2}{*}{Clean pan} & Ours  & \textbf{9/10} & \textbf{6/10} & \textbf{5/10} & \textbf{7/10} & \textbf{4/10} & \textbf{7/10} & \textbf{8/10} & \textbf{4/10} & \textbf{5/10} \\

\multirow{2}{*}{Uncover \& pour}  & AnyTeleop+ & 2/10 & 0/10 & 1/10 & 3/10 & \textbf{2/10} & 2/10 & 5/10 & 4/10 & 3/10 \\
& Ours & \textbf{8/10} & \textbf{4/10} & \textbf{8/10} & \textbf{6/10}  & \textbf{2/10} & \textbf{3/10} & \textbf{7/10} & \textbf{5/10} & \textbf{7/10} \\




\bottomrule
\end{tabular}
}
\end{small}
\caption{\textbf{Imitation Learning for Real-World Tasks.} The Success Rate (SR) is based on 10 trials, with SR-SG indicating success rates for spatial generalization and SR-U for unseen objects.}
\label{tab:imitation_learning}
\vspace{-20pt}
\end{table}

\vspace{-3pt}
\subsection{Tactile Data as Policy Input}
\vspace{-5pt}
\noindent\textbf{Tactile Data Processing.}
FSR sensors capture the pressure on the robot during object gripping. These signals can also be processed to calculate impulse by differentiating force over time. For tactile learning, touch signals are represented as vectors, encoded using MLPs, or visualized as point sets for visual embedding learning, akin to DexPoint~\citep{qin2023dexpoint} (see Appendix for details).

\noindent\textbf{Results and Discussion.}
As shown in Tab.~\ref{tab:touch_analysis}, utilizing tactile data does not necessarily enhance outcomes, potentially because vision alone suffice for these tasks. Tactile feedback only activates upon the contact between hands and objects, thus not aiding in the object's identification or localization. Moreover, using force data delivers slightly inferior performance compared to impulse data, possibly due to the zero-point drift in tactile readings over time. This drift, likely caused by the deformation of viscous materials used in sensor embedding, can persistently alter force reading during tasks, leading to unstable performance. Impulse, by contrast, is more robust against such variations.

\begin{table*}[th]
\vspace{-5pt}
\begin{minipage}[5cm]{0.45\linewidth}    
\centering
\begin{small}
\setlength\tabcolsep{5pt}
\scalebox{1.0}{
\begin{tabular}{lccc}
\toprule
\multirow{2}{*}{Task} & \multicolumn{3}{c}{DP3~\citep{Ze2024DP3}} \\
\cmidrule{2-3}
 &  SR-subtask & SR & \\
 \midrule

Wipe glass & 7/10 & 7/10 \\

Sweep floor & 8/10 & 4/10 \\

Prepare coffee & 7/10 & 4/10 \\

\bottomrule
\end{tabular}
}
\caption{\textbf{DP3 Results for Long-Horizon Tasks.} SR-subtask is subtask success rate.}
\label{tab:imitation_learning_lh}
\end{small}
\end{minipage}
\hfill
\begin{minipage}[5cm]{0.51\linewidth}
\centering
\begin{small}
\setlength\tabcolsep{4pt}
\scalebox{0.93}{
\begin{tabular}{llccc}
\toprule
Signal & Representation  & Clean pan & Uncover \& pour \\
\midrule
/ & / & \textbf{8/10} & 7/10\\
force & vector & 7/10 & 4/10 \\
impulse & vector & \textbf{8/10} & 7/10 \\
force & point cloud & 7/10 & 6/10 \\
impulse & point cloud & 7/10 & \textbf{8/10} \\
\bottomrule
\end{tabular}
}
\caption{\textbf{Ablation of Touch as DP3 Policy Input.} Success rates are reported for two bimanual tasks.}
\label{tab:touch_analysis}
\end{small}
\end{minipage}
\vspace{-10pt}
\end{table*}

\vspace{-3pt}
\section{Conclusion and Limitation}\label{sec:conclusion}
\vspace{-5pt}
\textbf{Limitation.}
Our system has two main limitations: (i) Vision Pro's hand tracking is inaccurate when fingers are self-occluded, causing jerky control commands. This issue could be mitigated by fusing data from wearable devices with Vision Pro. (ii) Subtle changes in haptic feedback are difficult to sense when the robot touches an object lightly. In future work, we will consider using actuators with larger contact areas and better sensitivity, such as piezoelectric actuators, to address this problem.

\textbf{Conclusion.}
In this paper, we introduce Bunny-VisionPro, a bimanual dexterous teleoperation system for real-time robotic control. Our system utilizes Vision Pro to track hand poses and convert them into control commands, addressing critical challenges including loop joint hand retargeting, collision avoidance, and singularity to ensure accurate, safe, and responsive operations. Moreover, we develop a haptic feedback device that delivers tactile sensations to the operator, thus enhancing control accuracy and fostering a more immersive experience.


\acknowledgments{We extend our sincere thanks to An-Chieh Cheng, Luobin Wang, Jiarui Xu, and Xinyu Zhang for their  efforts in testing and evaluating the teleoperation system.}

\clearpage
\bibliography{main}  

\clearpage
\appendix

\section*{Appendix Outline}\label{sec:appendix}

In these supplementary materials, we provide additional details for our Bunny-VisionPro teleoperation system, including:
\begin{itemize}
    \item [Sec.~\ref{sec:app_bimanual_init}] Bimanual teleoperation initialization modes.
    \item [Sec.~\ref{sec:app_sphere_model}] Sphere Modeling of Robot Links in Collision Checking.
    \item [Sec.~\ref{sec:app_task_suits}] Task suits design and demonstration details.
    \item [Sec.~\ref{sec:app_haptics}] Haptic feedback implementation.
    \item [Sec.~\ref{sec:app_tactile}] Tactile data processing for Imitation Learning.
    \item [Sec.~\ref{sec:app_policy_impl}] Imitation learning implementation.
\end{itemize}

\section{Bimanual Teleoperation Initialization}\label{sec:app_bimanual_init}
Teleoperation systems facilitate the mapping of human movements to robotic actions. For example, moving the right hand forward by 0.1 meters should result in an equivalent movement of the robot's right end effector. To ensure precise translation of human motions into robotic actions in a three-dimensional space, it is essential to define and synchronize coordinate systems for both the human operator and the robot during an initialization step. This phase involves establishing a three-dimensional framework, known as the initial human frame and the initial robot frame, for both systems. Movements made by the human operator are then measured relative to the initial human frame, and the robot replicates these movements within its own initial robot frame.

Bimanual teleoperation requires careful consideration of the relative positioning of the operator's two hands, which is critical for tasks that demand intricate dual-hand coordination. Given that the spacing between a robot's hands can differ from that of a human due to the specific hardware configuration, designing adaptable initialization modes that can dynamically align the robot's hands with those of the operator is vital. To address diverse operational requirements, we have developed several alignment modes:

\begin{itemize}
    \item  \texttt{ALIGN\_SEPARATELY}: Each arm is treated as an independent unit, which is ideal for tasks that require distinct arm movements.
    \item \texttt{ALIGN\_CENTER}: This mode calculates a midpoint between the two robot end effectors and aligns it with the central point between the human operator's hands.
    \item \texttt{ALIGN\_LEFT} and \texttt{ALIGN\_RIGHT}: These modes focus on aligning the robot's movements with either the left or the right hand of the human operator, which is beneficial for tasks that prioritize activity on one side.
\end{itemize}

Our system provides users with the flexibility to select from these modes, offering a range of options to suit specific task demands.

\section{Sphere Modeling in Collision Checking}\label{sec:app_sphere_model}
\begin{figure}
    \centering
    \includegraphics[width=\linewidth]{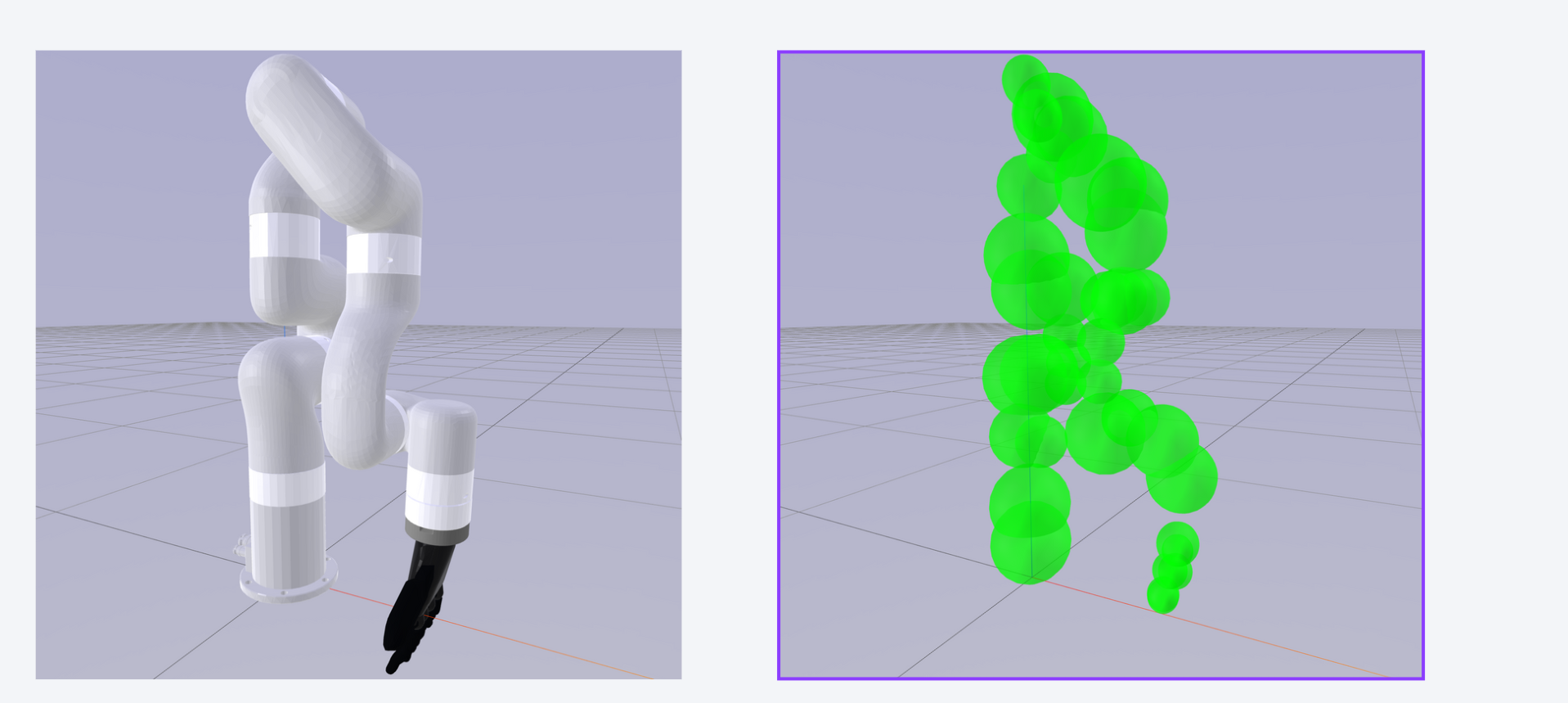}
    \caption{\textbf{Sphere Modeling of Robot Arm Links} for efficient and differentiable collision checking and avoidance.}
    \label{fig:app_sphere_model}
\end{figure}

To enable fast and differentiable collision avoidance during real-time teleoperation, we represent each link of the robot arm with multiple spheres of varying sizes, as detailed in Sec. 3.3 of the main paper. Fig.~\ref{fig:app_sphere_model} depicts these modeled spheres. We expedite the collision detection process by ignoring collisions between spheres within the same link or between a parent link and its direct child. This approach allows us to efficiently compute the gradient necessary for optimizing motion control in collision avoidance scenarios.

\section{Task Suits Design}\label{sec:app_task_suits}

\begin{figure}
    \centering
    \includegraphics[width=\linewidth]{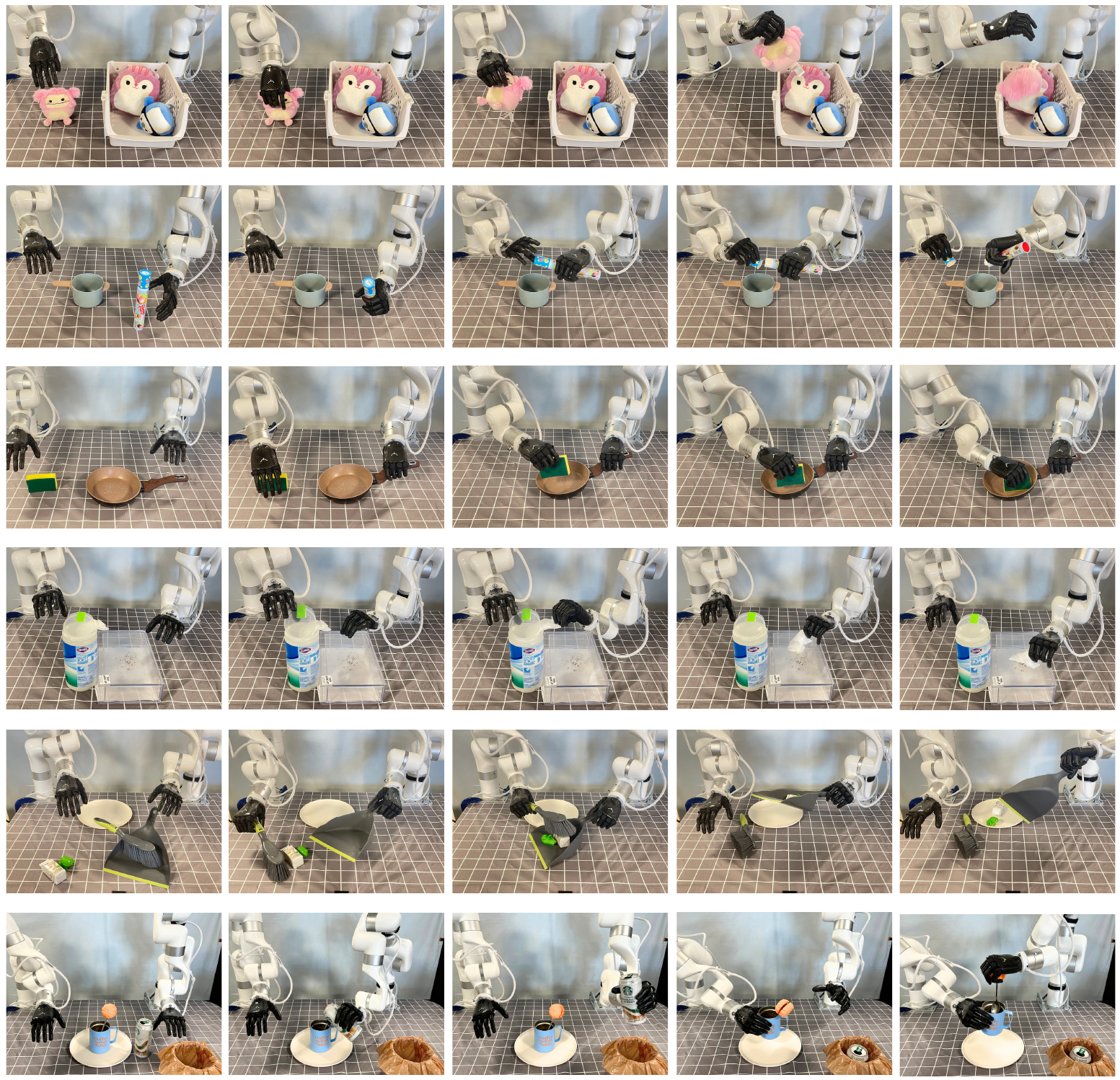}
    \caption{\textbf{Task Pipeline for Teleoperation and Imitation Learning.} We design six tasks including three short-horizon ones (top three rows) -- \textit{grasping toy}, \textit{uncovering and pouring} and \textit{cleaning pan} -- and three long-horizon ones (bottom three rows) -- \textit{wiping glass}, \textit{sweeping floor} and \textit{preparing coffee}.}
    \label{fig:app_task_pipeline}
\end{figure}

\begin{figure}
    \centering
    \includegraphics[width=\linewidth]{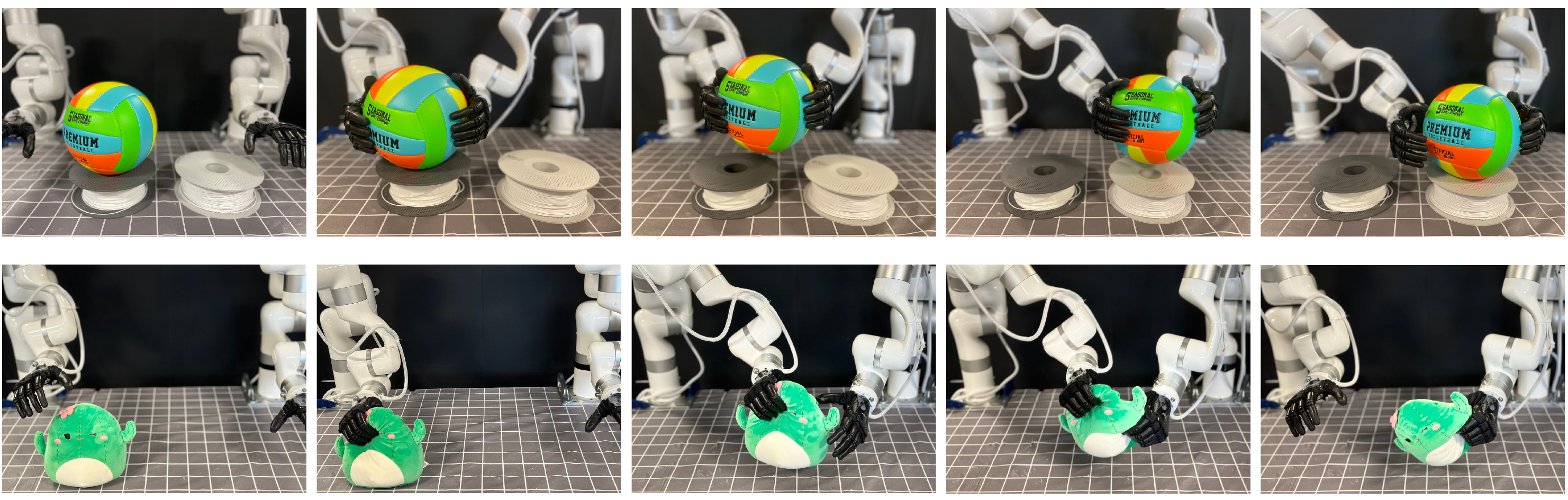}
    \caption{\textbf{Task Pipeline for Haptic User Study.} We design two tasks: \textit{ball moving} and \textit{toy handover}. The former requires precise and careful control due to the minimal deformability of the balls, while the latter involves manipulating soft and deformable toys.}
    \label{fig:app_user_study_tasks}
\end{figure}

\textbf{Task Suits Design.} 
In our main paper, teleoperation and imitation learning tasks are illustrated in Figure 1. Here, we provide the detailed task pipeline, depicted in Figure~\ref{fig:app_task_pipeline}. 
For the user study of haptics, we design \textit{toy handover} and  \textit{ball moving} tasks, which are illustrated in Fig.~\ref{fig:app_user_study_tasks}.

\textbf{Demonstration Details.} On the camera side, we collect images and processed point cloud data at approximately 30 Hz. Concurrently, the robots' proprioceptive data is recorded at over 100 Hz. These two data streams are timestamp-aligned via interpolation and downsampled to 10 Hz for imitation training. For short-horizon tasks, we gather 50 demonstrations per task; for long-horizon tasks, 30 demonstrations per task are collected.

\section{Haptic Feedback Implementation}\label{sec:app_haptics}

Our implementation of haptic feedback can be divided into two aspects: server-side signal processing and Arduino-side motor control.

\textbf{Server-side signal processing.} As outlined in Algorithm~\ref{alg:haptic_server}, the server side involves several stages of signal manipulation. Initially, tactile readings are calibrated, filtered, and normalized (refer to Section 3.4 for details). Subsequently, these processed signals are converted into valid PWM values and transmitted to the ELEGOO UNO board. This ensures that the tactile data are accurately prepared for motor activation, facilitating precise control over haptic feedback.

\textbf{Arduino-side motor control.} Algorithm~\ref{alg:haptic_board} details the pseudo-code for motor control on the Arduino side. This component of the system is responsible for parsing the PWM values received via serial communication from the server. Once parsed, these values are directly mapped to the PWM pins on the board, which in turn control the ERM motors. This arrangement enables the Arduino to dynamically adjust the intensity of the haptic feedback based on the input from the server, ensuring that the haptic responses are both timely and contextually appropriate.

\begin{algorithm}[H]
\caption{Server-side tactile signal processing}
\begin{algorithmic}[1]
\State \textbf{Initialize} serial connection with a specified baud rate
\newline

\While{$\text{True}$} \Comment{Until manually stopped}
    \State $V \gets \text{Get tactile sensor values from robot hand}$
    \State $\widehat V \gets \text{Calibrate and smoothen the tactile readings}$
    \State $T \gets \text{Set ERM activation threshold based on } \widehat V$
    \State $V_{\text{norm}} \gets \text{clip}(\left\lfloor \frac{(\widehat V - T) \times (255 - 0)}{\max(\widehat V) - T} \right\rfloor$, 0, 255) \Comment{Normalize $V$ and clip $V$ to [0, 255]}
    \State $\text{Send } V_{\text{norm}} \text{ to serial port as PWM values}$

\EndWhile
\end{algorithmic}
\label{alg:haptic_server}
\end{algorithm}

\begin{algorithm}[H]
\caption{Arduino-side motor controlling}
\begin{algorithmic}[1]
\State \textbf{Define} $numMotors$ \Comment{Total number of motors}
\State \textbf{Define} $motorPins[1 \ldots numMotors]$ \Comment{Array of PWM pins for each motor}
\newline

\Function{setup}{}
    \For{$i \gets 1$ \textbf{to} $numMotors$}
        \State $pinMode(motorPins[i], \text{OUTPUT})$ \Comment{Set each motor pin as an output}
    \EndFor
    \State \textbf{Initialize} serial communication with a specified baud rate
\EndFunction
\newline

\Function{loop}{}
    \If{$\text{Serial.available()} \geq numMotors \times \text{sizeof}(PWM\text{ value})$}
        \State $pwmValues[1 \ldots numMotors] \gets \text{Parse PWM values from Serial}$
        \For{$i \gets 1$ \textbf{to} $numMotors$}
            \State $\text{analogWrite}(motorPins[i], pwmValues[i])$ \Comment{Write PWM values to motor pins}
        \EndFor
    \EndIf
\EndFunction

\end{algorithmic}
\label{alg:haptic_board}
\end{algorithm}

\section{Tactile Data Processing}\label{sec:app_tactile}
We integrate tactile signals from the fingertips of the Ability hand into our imitation learning framework to evaluate its effectiveness across various conditions. On the \textbf{data acquisition} side, raw data from FSR sensors, which measure the pressure exerted on the robot during object manipulation, undergo zero-drift calibration and low-pass filtering similar to haptic feedback processing. Additionally, these signals can be processed to calculate force impulse by differentiating force over time. For tactile \textbf{data representation}, touch signals are effectively expressed as vectors, treated as components of the robot's state, and encoded using Multi-layer Perceptrons (MLPs). Furthermore, since touch indicates contact points between the robot and an object, these points are visualized as imaginary point sets. These are then concatenated with point cloud data from cameras to enhance visual embedding in the visual observation space for the imitation learning policy. To highlight active signals, we append a boolean dimension to indicate when tactile signals exceed predefined thresholds: 500 for force data and 50 for impulse data. This representation is inspired and adapted from DexPoint~\citep{qin2023dexpoint}.

\section{Imitation Learning Implementation}\label{sec:app_policy_impl}
We implement ACT~\citep{zhao2023learning}, Diffusion Policy~\citep{chi2023diffusionpolicy} and 3D Diffusion Policy~\citep{Ze2024DP3} to rigorously evaluate the demonstration quality collected by our system. To enhance performance assessment, we design generalization experiments focusing on the spatial locations of objects and interaction with unseen objects for each task. Details are further elaborated below.

\textbf{Learning Algorithms.} For Diffusion Policy, we employ multi-view images as visual inputs and establish a horizon of 8, consisting of 2 observation steps and 6 action steps. The model is trained for 300 epochs with a batch size of 64. For 3D Diffusion Policy, we utilize multi-view point clouds as inputs and maintain the same horizon settings as the Diffusion Policy. This model is trained for 500 epochs with a batch size of 64. For ACT, multi-view images are used, the chunk size is set as 20, and the training extends 3000 epochs.

\begin{figure}
    \centering
    \includegraphics[width=\linewidth]{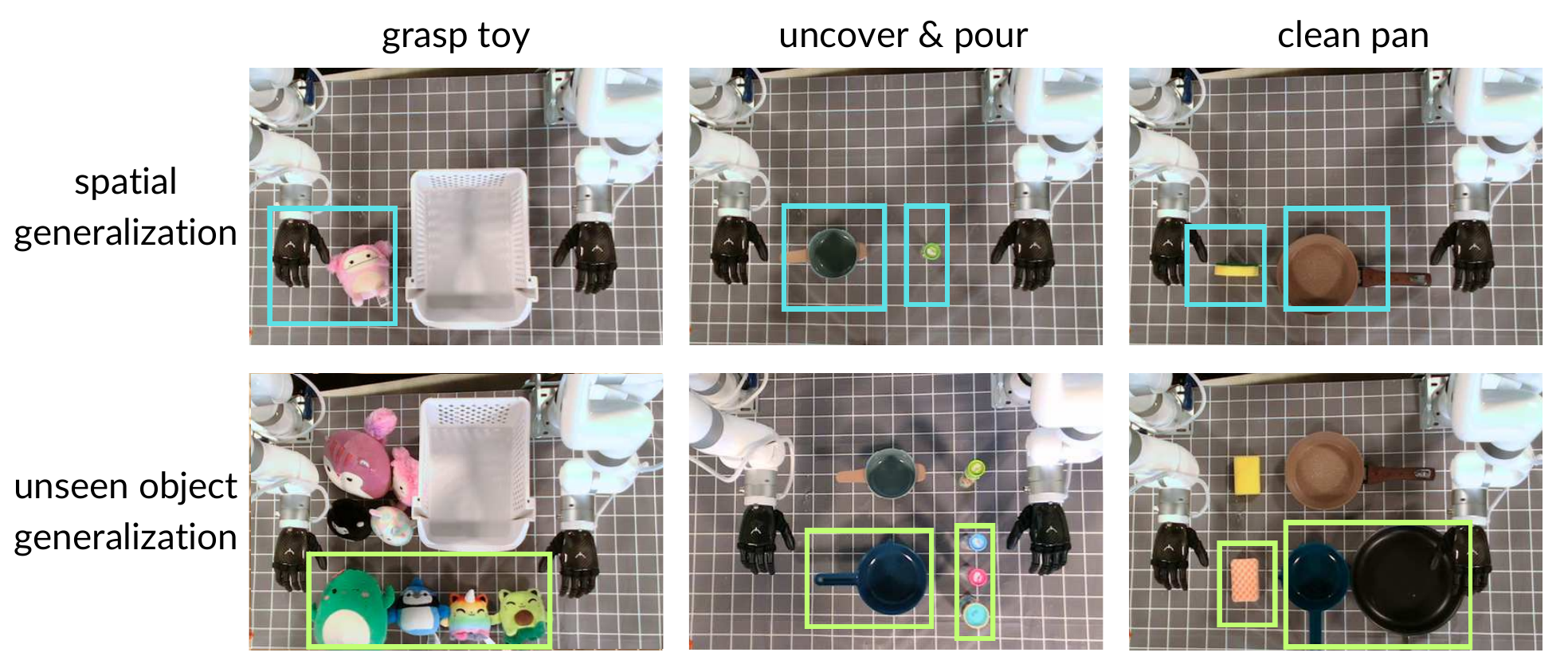}
    \caption{\textbf{Generalization Design of Imitation Learning.} \fcolorbox{figblue}{white}{\textcolor{figblue}{Blue boxes}} represent the spatial generalization range for objects, whereas \fcolorbox{figgreen}{white}{\textcolor{figgreen}{green boxes}} identify unseen objects that were not present during the collection of demonstrations.}
    \label{fig:app_il_generalization}
\end{figure}

\textbf{Generalization Evaluation Design.}
For generalization experiments involving different spatial locations and novel objects not present in the demonstrations, we detail the design for each task in Fig.~\ref{fig:app_il_generalization}. The spatial range encompasses a large working space, while the unseen objects vary in size, color, and shape, providing a comprehensive evaluation of the robustness of policies learned from our demonstrations.

\end{document}